# SLCNN: Sentence-Level Convolutional Neural Network for Text Classification

*Ali Jarrahi[α*], Leila Safari[α], Ramin Mousa[α]*

[α]*University of Zanjan, Zanjan , Iran*

A B S T R A C T

Text classification is a fundamental task in natural language processing (NLP). Several recent studies show the success of deep learning on text processing. Convolutional neural network (CNN), as a popular deep learning model, has shown remarkable success in the task of text classification. In this paper, new baseline models have been studied for text classification using CNN. In these models, documents are fed to the network as a three-dimensional tensor representation to provide sentence-level analysis. Applying such a method enables the models to take advantage of the positional information of the sentences in the text. Besides, analysing adjacent sentences allows extracting additional features. The proposed models have been compared with the state-of-the-art models using several datasets. The results have shown that the proposed models have better performance, particularly in the longer documents.

## 1. Introduction

In recent years, the production of unstructured texts (documents) has grown exponentially. Unstructured texts can be found everywhere, e.g., emails, social media, chat conversations, comments and websites. Although text data can be a rich source of information, it is hard to extract value from this type of unstructured data.

Text classification is a fundamental task in natural language processing (NLP). The task is the process of assigning a class label from a set of predefined classes to a given text according to its content, and has many applications such as sentiment analysis (Pang, Lee, and Vaithyanathan 2002) , spam detection (Jindal and Liu 2007) and topic categorization (Blei 2012).

Text classification can be done manually or automatically. Despite the manual method is more accurate, it is very costly and time consuming. Therefore, to provide scalability, several machine learning, NLP and other techniques are used for automatic text classification.

Supervised learning is a machine learning task of learning a function (classifier) using pre-labeled samples as a training dataset (Russell and Norvig 2010). A key step in supervised learning is feature extraction. Traditional machine learning methods represent text with hand-crafted methods, e.g., n-grams (Wang and Manning 2012). Recently, deep learning methods have been used for automatic feature extraction, including convolutional neural networks (CNNs) (LeCun et al. 1989), recurrent neural networks (RNNs) (Lipton 2015) and particularly long short-term memory (LSTM) (Hochreiter and Schmidhuber 1997).

In this paper, we present a new baseline model for text classification using CNN. In this model, documents are fed to the network as a three-dimensional tensor representation to provide sentence-level analysis.

The paper is structured as follows. The previous works have been summarized in the next section. The details of the proposed methods are described in section 3. We have evaluated our approach on several benchmark datasets. The experimental results are presented in section 4. Finally, the paper concludes with future research directions in section 5.

\* *Corresponding author.*
E-mail address: jarrahi@znu.ac.ir

## 2. Related works

Different approaches have been proposed for text classification. Initial approaches were based on the classical machine learning techniques, which followed two stages, i.e., extracting hand-crafted features and classifying the documents. Typical features include bag-of-words (BoW), n-grams, and their TF-IDF[*] (Zhang, Zhao, and LeCun 2015). Alternatively, several recent studies show the success of deep learning on text classification. As the neural networks receive their inputs numerically, word embeddings, e.g., word2vec (Mikolov et al. 2013) or GloVe (Pennington, Socher, and Manning 2014), are usually used to represent words as a numerical vectors by capturing the similarities/regularities between words.

There are variety of deep learning models for text classification. Due to the sequential nature of textual data, recurrent neural networks (RNN), including long short-term memory (LSTM) and gated recurrent units (GRU) (Cho et al. 2014) have been widely used in text processing. For example, in (Yogatama et al. 2017) authors examined generative and discriminative LSTM models for text classification. They found that although the generative models perform better than BoW, they have a higher asymptotic error rates than discriminative RNN-based models.

Another popular model is CNN, which originally invented for computer vision (LeCun et al. 1998). Subsequently, CNN models have been applied in NLP and have achieved excellent results (Collobert et al. 2011). Many researchers have worked on the effective use of CNNs in text classification since a single layer word-level CNN was successfully used in sentence classification with a pre-trained word embeddings (Kim 2014). The proposed method in (Zhang, Zhao, and LeCun 2015) was the first attempt to perform text classification entirely at the character-level, and reported competitive results. Their models use 70 characters by one-hot encoding, including 26 English letters, 10 digits, 33 other characters and the new line character. (Conneau et al. 2017) adopted very deep convolutional networks, i.e., ResNet (He et al. 2015), to the character-level text classification.

Some researchers tried to improve performance of the models by applying extra mechanisms. Attention is one of the most effective mechanism that selects significant information to achieve superior results (Ashish et al. 2017). Deep neural networks with attention mechanism can yield better results. Some of the remarkable examples include source-target attention and self-attention (Lin et al. 2017). Particularly, two-level attention mechanism, i.e., word attention and sentence attention, was developed on GRU by (Yang et al. 2016) for document classification. In (Wang, Huang, and Deng 2018), authors used dense connections with multi-scale feature attention in order to produce variable n-gram features. Since this paper aims to present a new baseline model, employing such mechanisms has been avoided.

## 3. Method

In this section, we describe the architecture of proposed Sentence-Level Convolutional Neural Network (SLCNN) for classifying the documents. The key idea of the model is that using positional information of each sentence in the document may improve the performance of the classifier. Furthermore, analysing adjacent sentences allows extracting some extra features, e.g., writing style features, which can be useful in some applications, such as spam review detection and fake news detection. Hence, we present two baseline models based on the CNN architecture for the text classification task. For this purpose, we introduce a three-dimensional representation of documents to enable sentence-level analysis. The pre-processing phase and the architecture of the SLCNN and its variant SLCNN+V are explained in the following subsections.

### 3-1. Pre-processing

During the pre-processing phase, the documents are cleaned by removing some unimportant characters, like the html tags and the punctuations. Then all words are normalized by converting to their lowercase forms. After that, as the most important step, each document is transformed into a three-dimensional tensor, illustrated in Figure 1. As shown in the figure, the sentences of the document form the first dimension of the tensor. In the same way, the words of the sentences shape the second dimension, while the third dimension represents the word vectors of the words. The pre-trained word embeddings, e.g., word2vec and GloVe, could be used for representing the word vectors.

Since, the input size of the network must be fixed, and according to different size of both the texts and the sentences, we consider two thresholds, one for the number of sentences in the documents, $T_d$, and another for the number of words in the sentences, $T_s$. The documents and the sentences longer than the thresholds would be cropped and shorter ones would be padded by zeros.

---

[*] Term Frequency–Inverse Document Frequency

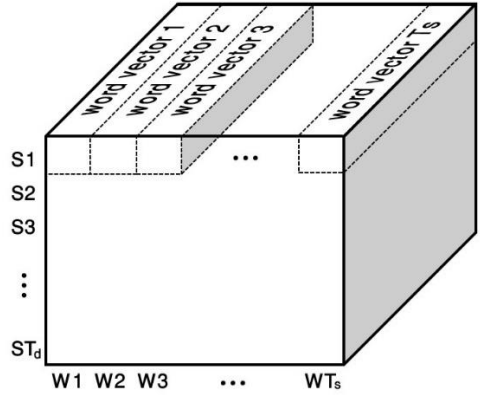

Figure 1- Shape of the converted documents.

After some statistical analysis on the datasets in our experiments, as well as considering the structure of the SLCNN, we chose $T_s$=46. In the same way, the threshold for the number of sentences in the documents is calculated by the following equation:

$$T_d = \lceil \mu + 1.5\, \sigma \rceil \qquad (1)$$

where μ is the average number of sentences in the documents, and $\sigma$ is the standard deviation. As a result, the outlier sizes are ignored to prevent model from constructing very large and sparse tensors. The relevant statistical data is provided in section 4.

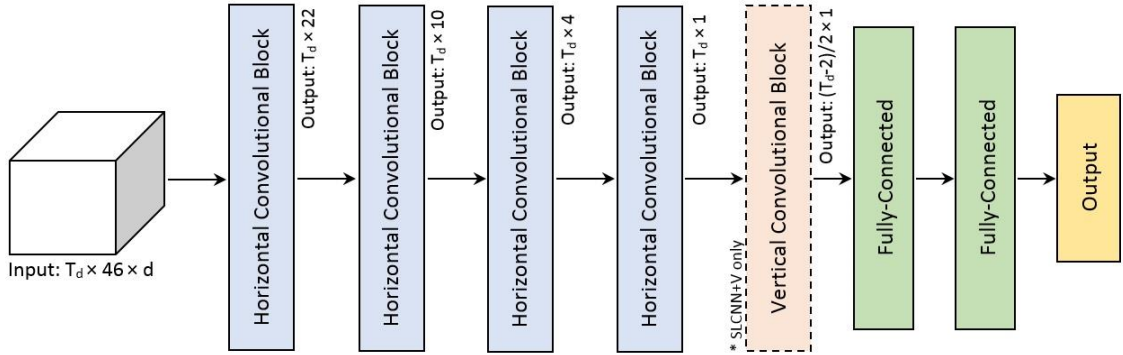

Figure 2- The architecture of the proposed models. The dashed block (VCB) is used only in SLCNN+V.

## 3-2. The Architecture

The architecture of the proposed models is illustrated in Figure 2. Overall, in the input layer, the documents are provided in the form of the 3D tensor, introduced in section 3-1. After that, using four horizontal convolutional blocks (HCB), one feature per filter is extracted for each sentence individually. In other words, one feature vector for each sentence is provided just before the fully-connected layers with the size equal to the number of filters. In this way, in addition to the word-level features, the positional information of the sentences is also used in the learning process. Moreover, as mentioned before, analysing of adjacent sentences can extract some useful features. For this purpose, the second model (SLCNN+V) is created by adding a vertical convolutional block (VCB) before fully-connected layers. Finally, there are two fully-connected (dense) layers which end to the output layer.

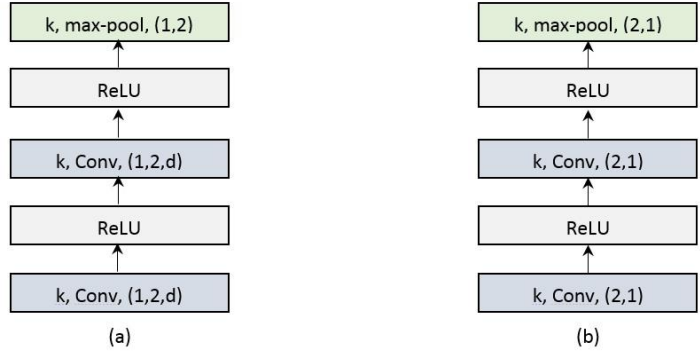

Figure 3- The convolutional blocks. k is the number of filters. (a) HCB and (b) VCB.

Looking at the details of the convolutional blocks, as shown in Figure 3, there are two sequential convolution layers, each one followed by a Rectified Linear Unit (ReLU) activation function, *f(x)= max (0, x)*. A convolution operation consists of a filter $w \in \mathbb{R}^{s \times t \times d}$, which is applied to each possible window of *s×t* features from its input *feature map, X,* to produce a new feature map by equation 3:

$$X = \begin{bmatrix} x_{1,1} & x_{1,2} & \cdots & x_{1,n} \\ x_{2,1} & x_{2,2} & \cdots & x_{2,n} \\ \vdots & \vdots & & \vdots \\ x_{m,1} & x_{m,2} & \cdots & x_{m,n} \end{bmatrix} \quad (2)$$

$$\tilde{x}_{i,j} = f(w \cdot x_{i,j:i+s-1,j+t-1} + b) \quad (3)$$

where $x_{i,j:y,z}$ is the concatenation of features within the specified interval, $b \in \mathbb{R}$ is a bias term and *f* is a non-linear function such as the ReLU. For the HCB, we consider *s=1* and *t=2*, and for the VCB *s=2* and *t=1*. It should be noted that, in the first convolution layer of the first HCB, *d* (the third dimension of the filters) is equal to the size of the word vectors, and in other cases *d=1*. At the end of the blocks, there is a max-pooling operation, with the pooling size = 2, that is applied over the generated intermediate feature map to select the maximum value from any two adjacent features as a more important feature. The new feature map is calculated by following equations:

$$\tilde{x}_{i,j} = \begin{cases} \max\{x_{i,2j-1}, x_{i,2j}\} & , for\ the\ HCB \\ \max\{x_{2i-1,j}, x_{2i,j}\} & , for\ the\ VCB \end{cases} \quad (4)$$

The process of extracting one feature from one filter was described. The model uses multiple filters to obtain multiple features. The final extracted features are passed to the fully-connected layers that end to a softmax output layer which is the probability distribution over labels. For regularization, a dropout module (Hinton et al. 2012) is employed after each fully-connected layer.

## 4. Experiments

### 4-1. Experimental settings

The Natural Language Toolkit (NLTK) was used in order to tokenize words and sentences. In the input layer, as mentioned before, pre-trained word-embeddings are used to convert the words into the corresponding word vectors. We used 100-dimensional GloVe in our experiments. Out-Of-Vocabulary (OOV) words were initialized from a uniform distribution with range [-0.01, 0.01]. We set number of filters to 128 for all the convolutional blocks. Also, we considered two different sizes for fully-connected layers, shown in Table 1. Both the dropout rates were set to 0.5. The model's parameters were trained by the Adam Optimizer (Kingma and Ba 2014), with the initial learning rate of 0.001. The model has been implemented using Keras and run for 50 epochs.

Table 1- Fully-connected layers in our experiments.

| Layers | Small | Large |
|---|---|---|
| Fully-connected 1 | 512 | 1024 |
| Fully-connected 2 | 512 | 1024 |
| Output | Depends on the problem | |

## 4-2. Benchmark Datasets

We utilized six datasets covering different classification tasks compiled by (Zhang, Zhao, and LeCun 2015). General specifications are presented in Table 2. All data are evenly distributed across class labels. AG and DBPedia are news and ontology classification datasets, respectively. Yelp and Amazon are sentiment classification datasets, where '.P' (Polarity) in the dataset names indicates that the labels are binary while '.F' (Full) means that the labels refer to the number of stars.

Table 2- Datasets in our experiments.

| Datasets | AG News | DBPedia | Yelp.P | Yelp.F | Amazon.P | Amazon.F |
|---|---|---|---|---|---|---|
| # of training samples | 120k | 560k | 560k | 650k | 3600k | 3000k |
| # of test samples | 7.6k | 70k | 38k | 50k | 400k | 650k |
| # of classes | 4 | 14 | 2 | 5 | 2 | 5 |

Some of the statistical information extracted from the datasets, after the pre-processing step, is summarized in Table 3. As presented in the table, by considering $T_s$=46, the proportions of cropped sentences are between 2 and 2.9 percent, that shows the length of sentences in the different datasets are almost similar. By contrast, the number of sentences of the documents in the different datasets are quite different. By utilizing Equation 1, $T_d$ for AG News, DBPedia, Amazon and Yelp are equal to 4, 6, 10 and 20 respectively. Also, the proportions of cropped documents, using relevant $T_d$, are 0.4, 3, 3.6 and 6 percent for AG News, Amazon, DBPedia and Yelp respectively, which means that the variance of the number of sentences in the documents of Yelp is greater than others.

Table 3- The statistical information of the datasets.

| Statistics | AG | DBPedia | Yelp.P | Yelp.F | Amazon.P | Amazon.F |
|---|---|---|---|---|---|---|
| # of sentences | 164k | 1505k | 5082k | 5958k | 18654k | 16986k |
| Cropped sentences (%) | 2 | 2.9 | 2.6 | 2.6 | 2.4 | 2.5 |
| Cropped documents (%) | 0.4 | 3.6 | 6 | 6 | 3.1 | 3 |
| Documents that contain cropped sentences (%) | 2.5 | 6.9 | 16.1 | 16.4 | 10.3 | 10.9 |
| # of sentences in the longest text | 15 | 25 | 141 | 151 | 85 | 99 |
| # of words in the longest sentence | 135 | 1302 | 1104 | 1175 | 522 | 520 |
| Vocab size | 62k | 786k | 283k | 311k | 1546k | 1464k |
| $T_d$ | 4 | 6 | 20 | 20 | 10 | 10 |
| # of trainable parameters in SLCNN *small* | 783k | 920k | 1831k | 1832k | 1176k | 1177k |
| # of trainable parameters in SLCNN *large* | 1835k | 2107k | 3930k | 3933k | 2619k | 2622k |
| # of trainable parameters in SLCNN+V *small* | 653k | 723k | 1176k | 1177k | 848k | 850k |
| # of trainable parameters in SLCNN+V *large* | 1508k | 1649k | 2554k | 2557k | 1899k | 1902k |
| Training time for a single epoch (s) | 10 | 51 | 150 | 170 | 510 | 440 |

## 4-3. Results

We compared our models with several popular base models, e.g., linear models (Zhang, Zhao, and LeCun 2015), RNN-based model, i.e., Discriminative-LSTM (Yogatama et al. 2017), and CNN-based models including classical word-level CNN (Kim 2014), character-level CNN (Zhang, Zhao, and LeCun 2015), very deep CNN (Conneau et al. 2017) and CNN with fastText embedding (Joulin et al. 2017). Since our aim was to provide new baseline models, and using other mechanisms, such as the attention, has been avoided, therefore such models have been excluded from the comparison. The results are listed in Table 4 based on accuracy. Overall, it can be seen that the proposed models have outperformed all the models in half of the datasets, DBPedia, Yelp.P and Yelp.F. Especially, the improvement is significant in Yelp datasets, i.e., around 2 percent in Yelp.P and around 5 percent in Yelp.F compared to character-level and word-level CNNs. In terms of Amazon datasets, the SLCNN+V was ranked third after VDCNN and character-level CNN with around 94 and 58.1 percent in Amazon.P and Amazon.F, respectively.

If we look at AG News, despite competitive results with other CNN models, n-grams and Discriminative-LSTM have achieved better results. One of the main reasons we can mention is the number of sentences in the documents. So that the proposed models perform better in documents with large number of sentences, i.e., Yelp. Another reason that hinders better performance in Amazon datasets is the very high vocabulary size (see Table 3), since we used the word embedding with just over 1M vocabularies in our experiments.

Table 4- Test accuracy (%) of all the models on the datasets. Results marked with * are reported in (Wang, Huang, and Deng 2018) and others are reprinted from the references.

| Models | | AG | DBPedia | Yelp.P | Yelp.F | Amazon.P | Amazon.F |
|---|---|---|---|---|---|---|---|
| Linear | Bag of Words (Zhang et al. 2015) | 88.81 | 96.61 | 92.24 | 57.99 | 90.40 | 54.64 |
| | n-grams (Zhang et al. 2015) | 92.04 | 98.63 | 95.64 | 56.26 | 92.02 | 54.27 |
| | n-grams TFIDF (Zhang et al. 2015) | **92.36** | 98.69 | 95.44 | 54.80 | 91.54 | 52.44 |
| CNN | Char-level CNN *small* (Zhang et al. 2015) | 84.35 | 98.02 | 93.47 | 59.16 | 94.50 | 59.47 |
| | Char-level CNN *large* (Zhang et al. 2015) | 87.18 | 98.27 | 94.11 | 60.38 | 94.49 | 58.69 |
| | VDCNN- 29 layers (Conneau et al. 2017) | 91.27 | 98.71 | 95.72 | 64.26 | **95.69** | **63.00** |
| | Word-level CNN (Kim 2014)* | 91.60 | 98.60 | 93.50 | 61.00 | - | 57.40 |
| | fastText (Joulin et al. 2017) | 91.50 | 98.10 | 93.80 | 60.40 | 91.20 | 55.80 |
| RNN | Discriminative-LSTM (Yogatama et al. 2017) | 92.10 | 98.70 | 92.60 | 59.60 | - | - |
| Ours | SLCNN *small* | 91.22 | 98.75 | 96.03 | **64.67** | 93.87 | 58.03 |
| | SLCNN *large* | 91.26 | **98.76** | 96.01 | 64.56 | 93.93 | 58.02 |
| | SLCNN+V *small* | 91.45 | 98.73 | **96.09** | 64.46 | 93.91 | 58.11 |
| | SLCNN+V *large* | 91.39 | **98.76** | 96.07 | 64.39 | 93.94 | 58.05 |

## 5. Conclusion and future works

This paper offers new baseline models for text classification using a sentence-level CNN. The key idea is representing the documents as a 3D tensor to enable the models to sentence-level analysis. The proposed models have been compared with the state-of-the-art models using several datasets. The results have shown that the proposed models have better performance, particularly in the longer documents.

As future works, the attention mechanism will be utilized in the proposed models in order to improve the overall performance. Also, we will work on sentence standardization. We believe that applying a standard form of sentences enables the proposed models to use compositional methods (with different 3D filters), due to the 3D structure of the input tensor.